\documentclass{article}

\usepackage{amsfonts, amssymb, amscd, amsmath, enumerate, verbatim, amsthm}

\usepackage[T1]{fontenc}
\usepackage{tikz}
\usepackage{algorithm,algorithmic}
\usepackage{multirow}

\newtheorem{definition}{Definition}[section]
\usepackage{hyperref}

\begin{document}
	
	\title{Learning in Function Spaces: An Unified Functional Analytic View of Supervised and Unsupervised Learning}
	\author{K. Lakshmanan \\ Department of Computer Science and Engineering \\ Indian Insititute of Technology (BHU), Varanasi 221005.\\ Email: lakshmanank.cse@iitbhu.ac.in}
	\date{}
	
	\maketitle
	\begin{abstract}
		
		Many machine learning algorithms can be interpreted as procedures for
		estimating functions defined on the data distribution.
		In this paper we present a conceptual framework that formulates a wide
		range of learning problems as variational optimization over function
		spaces induced by the data distribution.
		Within this framework the data distribution defines operators that
		capture structural properties of the data, such as similarity relations
		or statistical dependencies. Learning algorithms can then be viewed as
		estimating functions expressed in bases determined by these operators.
		
		This perspective provides a unified way to interpret several learning
		paradigms. In supervised learning the objective functional is defined
		using labeled data and typically corresponds to minimizing prediction
		risk, whereas unsupervised learning relies on structural properties of
		the input distribution and leads to objectives based on similarity or
		smoothness constraints. From this viewpoint, the distinction between
		learning paradigms arises primarily from the choice of the functional
		being optimized rather than from the underlying function space.
		
		We illustrate this framework by discussing connections with kernel
		methods, spectral clustering, and manifold learning, highlighting how
		operators induced by data distributions naturally define function
		representations used by learning algorithms. The goal of this work is
		not to introduce a new algorithm but to provide a conceptual framework
		that clarifies the role of function spaces and operators in modern
		machine learning.
	\end{abstract}


	\section{Introduction}
	
	Many machine learning algorithms can be interpreted as procedures for selecting functions in data-dependent function spaces. They are typically presented as optimization
	procedures over finite-dimensional parameter spaces. Examples include
	linear models, kernel methods, and neural networks, where learning
	amounts to adjusting parameters so that a model fits the observed data.
	Although this algorithmic viewpoint has proven extremely successful,
	it often obscures a deeper mathematical structure underlying many
	learning methods.
	
	We argue from a mathematical perspective that many machine learning problems can be
	interpreted as the task of selecting a function from an appropriate
	function space that best explains the data. Classical examples include
	kernel methods, where functions are selected from reproducing kernel
	Hilbert spaces, and spectral methods, where functions arise as
	eigenfunctions of operators derived from the data. These viewpoints
	suggest that a large class of learning algorithms can be understood
	within a common functional-analytic framework.
	
	In this paper we explore a conceptual formulation of machine learning
	based on optimization over function spaces. In this perspective, the
	data distribution induces operators that encode structural properties
	of the data such as similarity, geometry, and statistical dependence.
	Learning algorithms can then be viewed as procedures that estimate
	functions expressed in bases determined by these operators. In this framework we argue that supervised and unsupervised learning can be interpreted as instances of the same functional optimization problem.
	
	This viewpoint provides a unified way to interpret several seemingly
	different learning paradigms. Supervised learning can be viewed as
	estimating functions under constraints provided by labeled data,
	whereas unsupervised learning determines functions using structural
	properties of the data distribution alone. Spectral methods and kernel
	methods naturally arise through the analysis of operators associated
	with the data distribution.
	
	The goal of this paper is not to introduce a new learning algorithm,
	but rather to provide a conceptual framework that highlights the role
	of function spaces and operators in machine learning. By emphasizing
	this perspective, we aim to clarify connections between different
	learning methods and provide a mathematical viewpoint that complements
	the more common geometric and optimization-based interpretations. And we argue that many machine learning algorithms can be interpreted as instances of variational optimization over function spaces induced by the data distribution.
	
	Several existing machine learning frameworks already implicitly rely on
	function space viewpoints. Kernel methods operate in reproducing kernel
	Hilbert spaces, spectral clustering analyzes eigenfunctions of graph
	Laplacian operators, and convolutional neural networks apply sequences
	of convolution operators to construct hierarchical feature
	representations. More recently, operator-based formulations have also
	appeared in areas such as kernel mean embeddings \cite{muandet2017kernel}, diffusion maps, and
	operator learning \cite{li2020fourier}.
	
	These developments suggest that many learning algorithms can be viewed
	through a common mathematical lens: the data distribution induces
	operators whose spectral properties define natural function bases.
	Learning tasks then correspond to selecting functions expressed in
	these bases. This perspective emphasizes the role of operators and
	function spaces as organizing principles underlying diverse machine
	learning methods. 
	Several existing machine learning frameworks already implicitly rely on
	function space viewpoints. Kernel methods operate in reproducing kernel
	Hilbert spaces defined by positive definite kernels
	\cite{scholkopf2002learning,berlinet2004reproducing}. Spectral learning
	algorithms analyze eigenfunctions of graph Laplacian operators derived
	from similarity graphs \cite{vonluxburg2007spectral}. Manifold learning
	methods such as diffusion maps construct operators that approximate
	diffusion processes on the data distribution
	\cite{coifman2006diffusion}. More recently, operator-based
	representations of probability distributions have been developed using
	kernel mean embeddings \cite{muandet2017kernel}, while modern deep
	learning architectures can be interpreted as compositions of nonlinear
	operators that construct hierarchical representations
	\cite{goodfellow2016deep}. In addition, recent work on neural operators
	has explored learning mappings between function spaces directly from
	data \cite{li2020fourier}. These developments suggest that many learning
	algorithms can be understood through a common mathematical framework
	based on function spaces and operators induced by the data
	distribution.

\begin{figure}[t]
		\centering
		\begin{tikzpicture}[node distance=2.6cm,>=stealth]
			
			\node (data) [draw,rectangle,rounded corners] {Data Distribution $P_X$};
			
			\node (op) [draw,rectangle,rounded corners,below of=data]
			{Induced Operator};
			
			\node (basis) [draw,rectangle,rounded corners,below of=op]
			{Spectral Basis};
			
			\node (task) [draw,rectangle,rounded corners,below of=basis]
			{Learning Task};
			
			\node (alg) [draw,rectangle,rounded corners,below of=task]
			{Learning Algorithm};
			
			\draw[->] (data) -- (op);
			\draw[->] (op) -- (basis);
			\draw[->] (basis) -- (task);
			\draw[->] (task) -- (alg);
			
		\end{tikzpicture}
		
		\caption{
			Conceptual view of learning in function spaces. The data distribution
			induces operators whose spectral properties define function
			representations. Learning algorithms estimate functions in these
			representations to solve prediction or discovery tasks.
		}
		\label{fig:fa_ml}
	\end{figure}
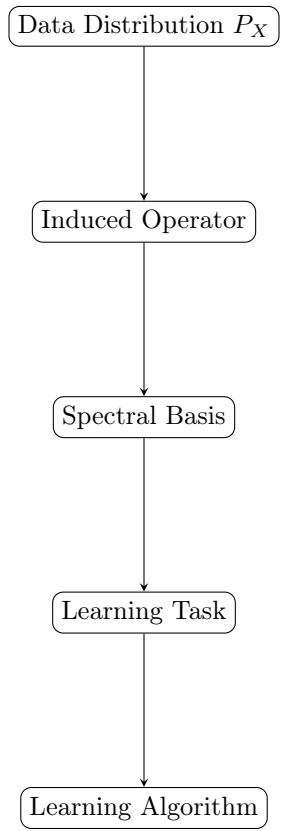

	Figure~\ref{fig:fa_ml} illustrates this conceptual view. The data
	distribution induces operators capturing structural properties of the
	data. These operators define function representations that are used by
	learning algorithms to estimate predictive or descriptive functions.
	
	The remainder of the paper is organized as follows. Section~2 introduces
	a general formulation of learning as optimization over function spaces.
	Section~3 describes operators induced by data distributions and their
	role in several machine learning methods. Sections~4 and~5 illustrate
	the framework using supervised learning and clustering. Section~6
	discusses connections with spectral and kernel methods. Finally,
	Section~7 summarizes the conceptual insights of this viewpoint and
	suggests directions for further exploration.

	\section{Learning in Function Spaces}
	
	\paragraph{Notation.}
	Let $X$ denote the input space and let $P_X$ be the probability
	distribution generating the data.
	Random variables are denoted by uppercase letters such as $X$ and $Y$,
	while their realizations are denoted by lowercase letters.
	We consider function spaces $\mathcal{H} \subset L^2(P_X)$ equipped
	with an inner product $\langle \cdot,\cdot \rangle$ and norm
	$\|\cdot\|_{\mathcal H}$.
	For a dataset $\{x_i\}_{i=1}^n$, functions defined on the data may also
	be represented as vectors $f=(f(x_1),\dots,f(x_n))$.
	Operators induced by the data distribution are denoted by uppercase
	letters such as $C$, $L$, or $T$.
	
	Many machine learning algorithms are typically described as procedures
	for optimizing parameters in a finite-dimensional model. Examples include
	linear models, neural networks, and kernel methods, where learning
	amounts to adjusting parameters so that the model fits observed data.
	While this parameter-centric viewpoint is convenient for computation,
	it often hides a deeper mathematical structure underlying many learning
	problems.
	
	A more general perspective is to view learning as the task of selecting
	a function from an appropriate function space. In this paper we adopt
	this viewpoint and interpret machine learning problems as optimization
	problems over spaces of functions.
	
	Let $X$ denote the input space and let $P_X$ be the probability
	distribution generating the data. Consider a function space
	$\mathcal{H} \subset L^2(P_X)$ equipped with an inner product
	$\langle \cdot,\cdot \rangle$ and corresponding norm $\|f\|_{\mathcal H}$.
	Typical examples of such spaces include reproducing kernel Hilbert
	spaces (RKHS \cite{berlinet2004reproducing}), Sobolev spaces, and other spaces commonly used in
	approximation theory.
	
	In many learning problems the goal is to determine a function
	\[
	f : X \rightarrow \mathbb{R}
	\]
	that captures some property of the data distribution.
	For example:
	
	\begin{itemize}
		\item in regression, $f(x)$ approximates the conditional expectation
		$\mathbb{E}[Y|X=x]$,
		\item in classification, the sign of $f(x)$ determines the predicted
		class label,
		\item in density estimation, $f(x)$ approximates the probability density
		of the data.
	\end{itemize}
	
	From this viewpoint learning can be interpreted as selecting a function
	\[
	f \in \mathcal H
	\]
	that minimizes a data–dependent functional.
	
	\begin{definition}
		Let $\mathcal{H}$ be a function space and let $P$ denote the information
		available about the data generating process.
		A learning problem consists of finding
		\[
		f^* = \arg\min_{f \in \mathcal H} J(f;P),
		\]
		where $J$ is a functional determined by the available data.
	\end{definition}
	
	The specific form of the functional $J$ depends on the learning task.
	For example, in supervised learning the functional depends on labeled
	samples, whereas in unsupervised learning it depends only on the
	structure of the data distribution. This formulation emphasizes that
	many learning problems share a common mathematical structure: they
	amount to selecting a function in a function space that optimizes a
	data-dependent objective.
	
	\paragraph{Example: Kernel Regression}
	
	Kernel regression provides a concrete example of the operator viewpoint.
	Given training samples $(x_i,y_i)$, kernel ridge regression estimates a
	function of the form
	
	\[
	f(x)=\sum_{i=1}^n \alpha_i k(x,x_i).
	\]
	
	The coefficients $\alpha_i$ are obtained by solving the regularized
	optimization problem
	
	\[
	\min_{\alpha} \sum_{i=1}^n
	\left(y_i-\sum_{j=1}^n \alpha_j k(x_i,x_j)\right)^2
	+ \lambda \alpha^T K \alpha .
	\]
	
	From the functional analytic perspective, this procedure estimates a
	function expressed in the spectral basis of the covariance operator
	associated with the kernel.
	
	\subsection{Variational Perspective}
	
	The formulation introduced above can also be interpreted from a
	variational perspective. Many learning problems can be expressed as
	variational optimization problems over function spaces.
	
	Let $\mathcal{H}$ be a function space and let $J : \mathcal{H} \to
	\mathbb{R}$ be a functional determined by the data. The learning
	problem can then be written as
	
	\[
	f^* = \arg\min_{f \in \mathcal{H}} J(f).
	\]
	
	From this viewpoint, machine learning algorithms can be interpreted as
	numerical procedures for solving variational optimization problems.
	The choice of the functional $J$ encodes the information available
	about the data, while the function space $\mathcal{H}$ determines the
	class of admissible solutions.
	
	This perspective connects machine learning with classical areas of
	mathematics such as variational methods, functional analysis, and
	operator theory. In particular, many learning algorithms can be viewed
	as approximations to variational problems defined on infinite-dimensional
	function spaces, with finite-dimensional parameterizations providing
	practical computational representations.
	
	\begin{center}
		\fbox{
			\begin{minipage}{0.9\linewidth}
				
				\textbf{Key Insight.}
				Many machine learning problems can be interpreted as selecting a
				function $f$ in a function space $\mathcal{H}$ that minimizes a
				data–dependent functional
				
				\[
				f^* = \arg\min_{f \in \mathcal H} J(f).
				\]
				
				Different learning paradigms correspond to different choices of the
				functional $J$:
				
				\begin{itemize}
					\item In \textbf{supervised learning}, $J$ depends on labeled data and
					corresponds to minimizing prediction risk.
					\item In \textbf{unsupervised learning}, $J$ depends only on structural
					properties of the data distribution, such as similarity or smoothness.
				\end{itemize}
				
				Thus supervised and unsupervised learning share a common functional
				framework, differing primarily in the objective functional rather than
				in the underlying function space.
				
			\end{minipage}
		}
	\end{center}
	
	\section{Operators Induced by Data Distributions}
	
	A useful way to analyze learning problems is through operators induced
	by the data distribution. Let $P_X$ denote the distribution of the input
	variable $X$ and let $\mathcal{H} \subset L^2(P_X)$ be a function space.
	Many machine learning methods can be interpreted in terms of linear
	operators acting on functions in $\mathcal{H}$.
	
	\subsection{Covariance Operators}
	
	Given a positive definite kernel $k : X \times X \to \mathbb{R}$,
	the distribution $P_X$ induces the covariance operator
	\[
	(Cf)(x) = \int k(x,x') f(x') \, dP_X(x').
	\]
	
	This operator plays a central role in kernel methods.
	Its eigenfunctions $\{\phi_j\}$ satisfy
	\[
	C\phi_j = \lambda_j \phi_j.
	\]
	
	These eigenfunctions form a natural basis for representing functions
	defined on the data distribution. Expansions in this basis appear in
	kernel PCA, kernel regression, and other kernel learning algorithms.
	
	\subsection{Graph Laplacian Operators}
	
	In many unsupervised learning problems the data are represented as a
	graph whose edge weights reflect similarity between samples.
	Let $W$ denote the similarity matrix and $D$ the degree matrix.
	The graph Laplacian is defined as
	\[
	L = D - W .
	\]
	
	The Laplacian acts as an operator on functions defined on the data
	points. Spectral clustering methods determine cluster structure by
	analyzing the eigenvectors of this operator. These eigenvectors
	approximate eigenfunctions of diffusion operators associated with the
	data distribution.
	
	\subsection{Conditional Expectation Operators}
	
	In supervised learning the relationship between inputs and labels can
	be expressed through the conditional expectation operator
	\[
	(Tf)(x) = \mathbb{E}[Y f(X) \mid X = x].
	\]
	
	Regression aims to approximate the function
	\[
	g(x) = \mathbb{E}[Y \mid X=x],
	\]
	which can be interpreted as a fixed point of such operators.
	Many learning algorithms implicitly estimate this function.
	
	\subsection{Convolution Operators}
	
	In signal processing and deep learning, convolutional neural networks
	can be interpreted in terms of convolution operators
	\[
	(Tf)(x) = \int h(x-x') f(x') dx'.
	\]
	
	Here $h$ represents a filter. The action of the operator extracts
	localized features of the signal or image. Stacking such operators
	creates hierarchical feature representations.
	
	\paragraph{Interpretation}
	
	These examples illustrate that many machine learning algorithms can be
	viewed as estimating functions expressed in bases determined by
	operators associated with the data distribution. Spectral properties
	of these operators encode important structural information about the
	data, including geometry, similarity relations, and statistical
	dependencies.

	\section{Supervised Learning}
	
	Supervised learning is one of the most widely studied problems in
	machine learning. In this setting we observe a collection of labeled
	samples
	\[
	(x_1,y_1), (x_2,y_2), \dots, (x_n,y_n),
	\]
	where $x_i \in X$ denotes the input and $y_i$ denotes the corresponding
	label or response variable. The objective is to learn a function
	\[
	f : X \rightarrow \mathbb{R}
	\]
	that predicts the label associated with a new input.
	
	From the functional analytic viewpoint introduced in the previous
	sections, supervised learning can be interpreted as selecting a function
	$f$ from a function space $\mathcal{H}$ that minimizes a suitable
	risk functional.
	
	\subsection{Risk Minimization}
	
	Let $(X,Y)$ denote the joint random variables generating the data.
	Given a loss function $L(\cdot,\cdot)$, the expected prediction error of
	a function $f$ is defined as
	
	\[
	R(f) = \mathbb{E}[L(f(X),Y)] .
	\]
	
	The goal of learning is to determine
	
	\[
	f^* = \arg\min_{f \in \mathcal{H}} R(f).
	\]
	
	In practice the distribution of $(X,Y)$ is unknown, so the risk is
	approximated using the empirical samples
	
	\[
	\hat{R}(f) = \frac{1}{n}\sum_{i=1}^n L(f(x_i),y_i).
	\]
	
	Regularization is typically introduced to control the complexity of the
	learned function. The resulting optimization problem takes the form
	
	\[
	f^* = \arg\min_{f \in \mathcal{H}}
	\left(
	\hat{R}(f) + \lambda \|f\|_{\mathcal H}^2
	\right).
	\]
	
	This formulation fits naturally within the general framework described
	in Section~2, where learning corresponds to minimizing a functional
	defined over a function space.
	
	\subsection{Regression}
	
	In regression problems the labels $y$ are real-valued.
	The optimal predictor under squared loss satisfies
	
	\[
	f^*(x) = \mathbb{E}[Y \mid X=x].
	\]
	
	Thus regression can be interpreted as estimating the conditional
	expectation function associated with the joint distribution of the
	data. Many regression algorithms—including kernel regression, Gaussian
	process regression, and neural network regression—can be viewed as
	approximating this function within a chosen function space.
	
	From the operator perspective discussed in Section~3, the conditional
	expectation operator defined by the joint distribution of $(X,Y)$
	plays a central role in determining the optimal prediction function.
	
	\subsection{Classification}
	
	In classification problems the labels belong to a finite set
	$\{1,\dots,K\}$. The goal is to determine a decision function that
	assigns an input $x$ to one of the classes.
	
	A common approach is to learn a real-valued function $f(x)$ and define
	the predicted class using its sign or maximum component. For example,
	in binary classification the decision rule is often written as
	
	\[
	\hat{y}(x) = \text{sign}(f(x)).
	\]
	
	Many classification algorithms—including support vector machines,
	logistic regression, and neural networks—can therefore be interpreted as
	learning functions in an appropriate function space.
	
	\subsection{Generalization}
	
	A central question in supervised learning concerns the ability of the
	learned function to generalize beyond the training data.
	Since the true risk
	
	\[
	R(f) = \mathbb{E}[L(f(X),Y)]
	\]
	
	depends on the unknown data distribution, practical algorithms minimize
	the empirical risk computed from the available samples.
	Understanding the relationship between empirical risk and true risk is
	a central topic in statistical learning theory \cite{vapnik1998statistical}.
	
	Within the functional framework described in this paper, generalization
	properties depend on the complexity of the function space
	$\mathcal{H}$. Measures such as Rademacher complexity, covering numbers,
	or other capacity measures are commonly used to quantify this
	complexity and to derive bounds on the difference between empirical and
	true risk.
	
	Thus, while learning algorithms compute functions using finite data,
	the underlying variational problem can be viewed as an optimization over
	a potentially infinite-dimensional function space whose complexity
	controls the generalization behavior of the learned model.
	
	\paragraph{Interpretation}
	
	From the functional analytic viewpoint, supervised learning can be
	interpreted as estimating a function that best approximates the
	relationship between inputs and labels under a specified loss
	functional. The choice of function space $\mathcal{H}$ determines the
	class of functions that the algorithm can represent, while the loss
	function determines the criterion used to evaluate predictions.
	
	This perspective highlights that the central object in supervised
	learning is the function being estimated, rather than the particular
	parametric form used to represent it.
	
	\section{Clustering and Unsupervised Learning}
	
	In unsupervised learning the data consist only of input samples
	\[
	x_1,x_2,\dots,x_n,
	\]
	without associated labels. The goal is to discover structural patterns
	in the data, such as clusters, low-dimensional manifolds, or latent
	representations.
	
	From the functional viewpoint introduced earlier, unsupervised learning
	can be interpreted as determining functions defined on the input space
	that reveal structural properties of the data distribution.
	
	\subsection{Clustering as a Function Estimation Problem}
	
	In clustering problems the objective is to partition the input space
	into groups of similar observations. One way to formalize this problem
	is to consider a function
	
	\[
	f : X \rightarrow \{1,\dots,K\},
	\]
	
	where $f(x)$ denotes the cluster label assigned to the point $x$.
	
	Under this formulation, clustering can be interpreted as selecting a
	function $f$ that minimizes a functional measuring the inconsistency of
	the clustering assignment with respect to a similarity structure among
	data points.
	
	For example, given a similarity measure $w(x_i,x_j)$ between points,
	a typical clustering objective takes the form
	
	\[
	J(f) = \sum_{i,j} w(x_i,x_j)\mathbf{1}(f(x_i) \neq f(x_j)).
	\]
	
	This functional penalizes assignments that place similar points in
	different clusters.
	
	\subsection{Example: k-Means Clustering}
	
	The well-known k-means algorithm can also be interpreted within the
	functional framework described in this paper. In k-means clustering,
	the goal is to partition the dataset into $K$ clusters by minimizing
	the within-cluster variance.
	
	Let $\mu_1,\dots,\mu_K$ denote cluster centers. The classical k-means
	objective can be written as
	
	\[
	J(\mu_1,\dots,\mu_K) =
	\sum_{i=1}^n \min_{k=1,\dots,K} \|x_i - \mu_k\|^2 .
	\]
	
	This objective can be reformulated using a cluster assignment function
	
	\[
	f : X \rightarrow \{1,\dots,K\},
	\]
	
	which assigns each data point to a cluster. The objective then becomes
	
	\[
	J(f,\mu) =
	\sum_{i=1}^n \|x_i - \mu_{f(x_i)}\|^2 .
	\]
	
	Under this formulation the clustering problem can be interpreted as
	selecting a function $f$ that minimizes a functional measuring the
	within-cluster distortion. The k-means algorithm alternates between
	estimating the function $f$ (cluster assignments) and updating the
	cluster centers $\mu_k$.
	
	From the perspective of this paper, k-means therefore fits naturally
	into the general learning framework
	
	\[
	f^* = \arg\min_{f \in \mathcal H} J(f),
	\]
	
	where the functional $J$ reflects the geometric structure of the
	dataset rather than prediction accuracy.
	
	\subsection{Graph-Based Clustering}
	
	Many modern clustering algorithms represent the dataset as a graph in
	which nodes correspond to data points and edges encode pairwise
	similarities.
	
	Let $W$ denote the similarity matrix with entries
	\[
	W_{ij} = w(x_i,x_j).
	\]
	
	The degree matrix $D$ is defined by
	\[
	D_{ii} = \sum_j W_{ij}.
	\]
	
	The graph Laplacian operator
	
	\[
	L = D - W
	\]
	
	plays a central role in spectral clustering methods \cite{vonluxburg2007spectral}.
	
	Functions defined on the nodes of the graph can be viewed as vectors
	$f = (f(x_1),\dots,f(x_n))$. The quadratic form
	
	\[
	f^T L f
	\]
	
	measures the smoothness of the function with respect to the similarity
	graph. Minimizing this quantity encourages similar data points to have
	similar function values.
	
	\subsection{Spectral Clustering}
	
	Spectral clustering algorithms compute eigenvectors of the graph
	Laplacian operator and use them to obtain a low-dimensional embedding
	of the data \cite{vonluxburg2007spectral}. Clusters are then identified in this embedding space using
	standard methods such as k-means.
	
	From the functional perspective, spectral clustering identifies
	functions associated with the smallest eigenvalues of the Laplacian
	operator. These functions capture large-scale structural properties of
	the data distribution and naturally reveal cluster boundaries.
	
	\paragraph{Interpretation}
	
	The operator viewpoint highlights that clustering algorithms can be
	interpreted as estimating functions that are smooth with respect to the
	similarity structure of the data. The Laplacian operator encodes this
	structure, and its eigenfunctions provide a natural basis for describing
	cluster assignments.
	
	Thus, unsupervised learning can also be interpreted as an optimization
	problem over functions defined on the data distribution, where the
	objective functional reflects structural properties of the dataset
	rather than prediction accuracy.
	
	\subsection{Relationship Between Supervised and Unsupervised Learning}
	
	The functional analytic perspective developed in this paper provides a
	useful way to compare supervised and unsupervised learning. Although
	these paradigms are often presented as fundamentally different, they
	share a common mathematical structure.
	
	In both cases the objective is to determine a function
	\[
	f : X \rightarrow \mathbb{R}
	\]
	belonging to a function space $\mathcal{H}$. The difference lies in the
	information used to define the functional being optimized.
	
	In supervised learning the functional depends on labeled data and is
	typically expressed through a risk functional such as
	
	\[
	J_{\text{sup}}(f) = \mathbb{E}[L(f(X),Y)] .
	\]
	
	The goal is to estimate a function that predicts the label associated
	with each input. In regression this corresponds to estimating the
	conditional expectation $\mathbb{E}[Y|X]$, while in classification the
	function determines the decision boundary between classes.
	
	In unsupervised learning the functional depends only on structural
	properties of the input distribution. For example, clustering
	algorithms often minimize objectives that encourage similar points to
	have similar function values, leading to functionals of the form
	
	\[
	J_{\text{unsup}}(f) = f^T L f ,
	\]
	
	where $L$ denotes a graph Laplacian operator constructed from the
	data.
	
	Thus both supervised and unsupervised learning can be interpreted as
	instances of the same general problem
	
	\[
	f^* = \arg\min_{f \in \mathcal{H}} J(f),
	\]
	
	with the difference lying in how the functional $J$ is defined.
	Supervised learning uses label information to define the objective,
	whereas unsupervised learning relies solely on structural properties of
	the data distribution.
	
	This viewpoint emphasizes that many learning problems share a common
	functional structure: they involve selecting functions in spaces
	defined by operators induced by the data distribution. The choice of
	objective functional determines whether the resulting learning problem
	is predictive (supervised) or descriptive (unsupervised).
	
	\section{Connections with Spectral and Kernel Methods}
	
	The functional analytic perspective described in the previous sections
	is closely related to several established areas of machine learning.
	In particular, kernel methods, spectral learning algorithms, and
	manifold learning techniques all implicitly rely on operators induced
	by the data distribution.
	
	\paragraph{Kernel Methods}
	
	Kernel methods provide one of the clearest examples of learning in
	function spaces. In these methods, functions are selected from a
	reproducing kernel Hilbert space (RKHS)  associated with a positive
	definite kernel $k(x,x')$ \cite{scholkopf2002learning}.
	
	The kernel defines an implicit feature map
	\[
	\phi(x) : X \rightarrow \mathcal{H},
	\]
	such that
	\[
	k(x,x') = \langle \phi(x), \phi(x') \rangle_{\mathcal H}.
	\]
	
	Learning algorithms such as kernel ridge regression and support vector
	machines determine functions of the form
	
	\[
	f(x) = \sum_{i=1}^{n} \alpha_i k(x,x_i),
	\]
	
	which correspond to linear functions in the feature space $\mathcal H$.
	From the operator viewpoint, the kernel induces a covariance operator
	whose spectral properties determine natural bases for representing
	functions on the data distribution.
	
	\paragraph{Spectral Learning}
	
	Spectral learning methods analyze the eigenstructure of operators
	constructed from data. A prominent example is spectral clustering,
	which uses eigenvectors of the graph Laplacian operator to reveal
	cluster structure.
	
	More generally, spectral methods identify functions associated with
	the dominant eigenvalues of operators derived from similarity
	relationships between data points. These eigenfunctions capture
	large-scale geometric properties of the data and provide useful
	low-dimensional representations.
	
	Such methods illustrate how operator eigenfunctions can serve as
	natural coordinate systems for learning problems.
	
	\paragraph{Manifold Learning}
	
	Manifold learning algorithms attempt to uncover low-dimensional
	structures embedded in high-dimensional datasets.
	Techniques such as diffusion maps, Laplacian eigenmaps, and related
	methods construct operators that approximate diffusion processes on
	the data manifold.
	
	The eigenfunctions of these operators provide coordinates that
	reflect the intrinsic geometry of the data.
	From the functional viewpoint developed in this paper,
	manifold learning can be interpreted as identifying basis functions
	associated with operators induced by the data distribution.
	
	\paragraph{Representation Learning.}
	
	Modern machine learning methods often focus on learning useful
	representations of data. In deep learning architectures \cite{goodfellow2016deep}, intermediate
	layers transform the input through a sequence of operators that
	progressively extract higher-level features. From the functional
	analytic viewpoint, these transformations can be interpreted as
	constructing new function representations adapted to the data
	distribution. The learned representations effectively define new
	function spaces in which prediction or decision functions are
	estimated. This perspective suggests that representation learning can
	be viewed as the process of discovering operators that generate useful
	function bases for downstream learning tasks.
	
	\paragraph{Operator Learning.}
	
	Recent work in machine learning has also explored the problem of
	learning operators directly from data \cite{li2020fourier}. Examples include neural operator
	models that learn mappings between function spaces in applications such
	as scientific computing and physical system modeling. These approaches
	emphasize learning operators rather than individual functions and
	highlight the importance of operator-based representations of data.
	The functional analytic framework described in this paper provides a
	natural conceptual setting for understanding such methods, since the
	learning problem can be interpreted in terms of estimating functions
	or operators defined on spaces induced by the data distribution.
	
	
	These connections highlight that many machine learning algorithms
	already operate implicitly within a functional analytic framework.
	Kernel methods select functions in RKHS spaces, spectral algorithms
	analyze eigenfunctions of data-dependent operators, and manifold
	learning methods approximate diffusion operators on the data manifold \cite{coifman2006diffusion}.
	
	The framework proposed in this paper provides a conceptual viewpoint
	that unifies these approaches by emphasizing the role of operators
	induced by the data distribution and the function spaces they define.
	
	\section{Discussion and Conclusion}
	
	\begin{figure}[t]\label{lpard}
		\centering
		\begin{tikzpicture}[node distance=3cm,>=stealth]
			
			\node (framework) [draw,rectangle,rounded corners,align=center]
			{Learning Framework\\
				$f^*=\arg\min_{f\in\mathcal H} J(f)$};
			
			\node (sup) [draw,rectangle,rounded corners,below left of=framework,align=center,xshift=-1cm]
			{Supervised Learning\\
				$J(f)=\mathbb{E}[L(f(X),Y)]$\\
				Prediction risk};
			
			\node (unsup) [draw,rectangle,rounded corners,below right of=framework,align=center,xshift=1cm]
			{Unsupervised Learning\\
				$J(f)=f^TLf$\\
				Similarity / smoothness};
			
			\node (cluster) [draw,rectangle,rounded corners,below of=framework,align=center,yshift=-3cm]
			{Clustering / k-means\\
				$J(f)=\sum \|x_i-\mu_{f(x_i)}\|^2$\\
				Distortion objective};
			
			\draw[->] (framework) -- (sup);
			\draw[->] (framework) -- (unsup);
			\draw[->] (framework) -- (cluster);
			
		\end{tikzpicture}
		
		\caption{
			Summary of the functional framework for machine learning.
			Different learning paradigms correspond to minimizing different
			functionals over a function space $\mathcal H$ induced by the data
			distribution.
		}
	\end{figure}
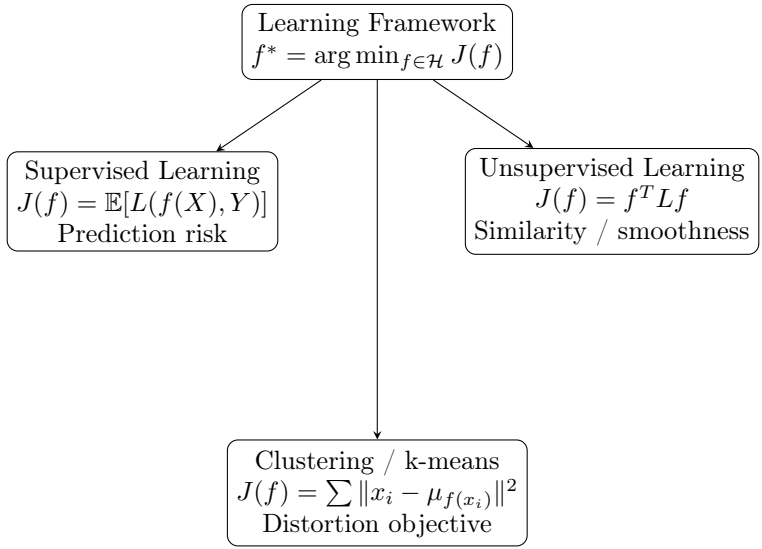
	
	In this paper we presented a conceptual framework for interpreting
	machine learning problems using ideas from functional analysis.
	The central viewpoint is that many learning problems can be formulated
	as optimization problems over function spaces. In this formulation the
	data distribution induces operators that encode structural properties
	of the data, while learning algorithms estimate functions expressed in
	bases determined by these operators.
	
	This perspective highlights a common mathematical structure underlying
	several learning paradigms. In supervised learning the objective
	functional is defined using labeled data and typically corresponds to
	minimizing a prediction risk. In contrast, unsupervised learning relies
	on structural properties of the data distribution, leading to objectives
	that encourage smoothness or consistency with respect to similarity
	relations among data points. From this viewpoint, the distinction
	between learning paradigms arises primarily from the choice of the
	functional being optimized rather than from the underlying function
	space. The relationships between these learning paradigms are summarized
	in Figure~\ref{lpard}, which illustrates how different machine learning tasks
	can be interpreted as minimizing distinct functionals over a common
	function space.
	
	The operator viewpoint also clarifies connections between several
	well-known machine learning methods. Kernel methods operate in
	reproducing kernel Hilbert spaces defined by covariance operators,
	spectral clustering analyzes eigenfunctions of graph Laplacian
	operators, and manifold learning methods estimate eigenfunctions of
	diffusion operators associated with the data distribution. These
	methods illustrate how operators induced by data can define natural
	function representations for learning tasks.
	
	Although the framework presented here is primarily conceptual, it
	suggests several possible directions for further investigation.
	For example, the operator viewpoint may provide a useful lens for
	analyzing representation learning methods, understanding the role of
	spectral structure in deep architectures, or developing learning
	algorithms that explicitly exploit operator-based representations of
	data.
	
	\[
	\text{Learning} =
	\arg\min_{f \in \mathcal H} J(f;P)
	\]
	
	More broadly, the functional analytic perspective emphasizes that the
	central object in many learning problems is the function being
	estimated rather than the particular parametric form used to represent
	it. By highlighting this viewpoint, we hope to encourage further
	connections between functional analysis and modern machine learning
	methods.
	
	\bibliographystyle{plain}
	\bibliography{mlfunc}
\end{document}